\documentclass[conference]{IEEEtran}
\IEEEoverridecommandlockouts
\usepackage{cite}
\usepackage{amsmath,amssymb,amsfonts,amsthm}
\usepackage{algorithmic}
\usepackage{graphicx}
\usepackage{textcomp}
\usepackage{xcolor}
\usepackage{url}
\usepackage{booktabs}
\usepackage{hyperref}
\usepackage[symbol]{footmisc}

\def\BibTeX{{\rm B\kern-.05em{\sc i\kern-.025em b}\kern-.08em
    T\kern-.1667em\lower.7ex\hbox{E}\kern-.125emX}}

\newtheorem{proposition}{Proposition}
\makeatletter
\def\th@proposition{\thm@headfont{\bfseries}\itshape}
\makeatother

\newcommand{\concat}{\mathop{\raisebox{-0.5ex}{\scalebox{1.5}{$\|$}}}}

\definecolor{darkgreen}{rgb}{0.0, 0.6, 0.0}
\definecolor{coolblue}{rgb}{0.0, 0.5, 0.75}

\hypersetup{
    colorlinks=true,
    linkcolor=red,
    urlcolor=coolblue,    
    citecolor=darkgreen,
}

\begin{document}

\title{Channel-Attentive Graph Neural Networks$^*$\thanks{$^*$Accepted to IEEE ICDM 2024: \url{https://ieeexplore.ieee.org/document/10884168}}}


\author{\IEEEauthorblockN{Tuğrul Hasan Karabulut, İnci M. Baytaş}
\IEEEauthorblockA{
\textit{Boğaziçi University}\\
Istanbul, Turkey \\
tugrul.karabulut@std.bogazici.edu.tr, inci.baytas@bogazici.edu.tr}
}
\maketitle

\begin{abstract}
Graph Neural Networks (GNNs) set the state-of-the-art in representation learning for graph-structured data. They are used in many domains, from online social networks to complex molecules. Most GNNs leverage the message-passing paradigm and achieve strong performances on various tasks. However, the message-passing mechanism used in most models suffers from over-smoothing as a GNN's depth increases. The over-smoothing degrades GNN's performance due to the increased similarity between the representations of unrelated nodes. This study proposes an adaptive channel-wise message-passing approach to alleviate the over-smoothing. The proposed model, Channel-Attentive GNN, learns how to attend to neighboring nodes and their feature channels. Thus, much diverse information can be transferred between nodes during message-passing. Experiments with widely used benchmark datasets show that the proposed model is more resistant to over-smoothing than baselines and achieves state-of-the-art performances for various graphs with strong heterophily. Our code is at \href{https://github.com/ALLab-Boun/CHAT-GNN}{https://github.com/ALLab-Boun/CHAT-GNN}.
\end{abstract}


\section{Introduction}
\label{intro}

Graph structure is ideal for modeling the interactions between the related entities. A wide variety of applications, including complex physical systems~\cite{physics}, recommender systems~\cite{recom2}, natural language processing~\cite{nlp1}, computer vision~\cite{vision1}, and protein structures~\cite{protein1}, benefit from modeling their data with graphs. One of the reasons behind the widespread use of graphs is the advances in representation learning with deep networks that offer solutions to learning task-specific representations from graphs. Particularly, Graph Neural Networks (GNNs) stand out as neural network architectures specialized in node-level and graph-level representation learning from graphs in an end-to-end fashion~\cite{gcn, gat, cheby, sage}.

GNNs constitute a family of neural network architectures that operate on graphs. The most recent state-of-the-art GNN architectures learn node representations by applying local operations on its neighborhood, also known as message-passing~\cite{gcn, sage, gat}. The message-passing mechanism aggregates the node's representation with the "messages" received from its neighborhood in various ways~\cite{gilmer}. The message-passing allows the model to encode information from further away nodes into the node's representation. One of the prominent techniques for aggregating local information is graph convolution. Graph Convolutional Networks (GCNs) perform the first-order approximation of spectral graph convolution, equivalent to aggregating the neighbor representations scaled by node degrees~\cite{gcn, cheby}. Moreover, Graph Attention Network (GAT) aims to learn each neighbor's contribution as attention weights to perform a more adaptive aggregation~\cite{gat}. On the other hand, previous studies, such as GraphSAGE, employ neighborhood sampling and handle the node and neighbor representations separately~\cite{sage}.
Although incorporating the information from further away nodes might be beneficial for specific tasks, when repeated in multiple layers, the message-passing mechanism could lead to an issue called over-smoothing~\cite{oversmoothing, gnn_exp, lowpass, measure_smooth}. The over-smoothing problem degrades the discriminability of the node representations. Using less number of message-passing layers could alleviate the problem. However, some topologies and tasks might require deeper propagation to exchange information between distant nodes. Therefore, there has been ongoing research to combat over-smoothing in recent years~\cite{measure_smooth, gcn2, aero, g2gnn}. One of the prominent approaches to alleviate over-smoothing is attention. The attention-based message-passing GNNs successfully achieve state-of-the-art performances by incorporating the self-attention mechanism into GNNs~\cite{gat, gatv2, gprgnn, dagnn, fagcn, aero}. The attention mechanism assigns importance scores to neighbors~\cite{gat, gatv2, fagcn, aero} or the aggregated messages from different hops~\cite{dagnn, gprgnn, aero}. The attention applied to edges helps in aggregating the neighbor representations by taking into account the generated importance scores. Meanwhile, the hop attention weights the representations from different layers. 

The attention layers in the literature provide more expressive aggregation mechanisms by attending to information from separate hops. However, the edge attention weights are scalars computed as a probability distribution over the neighborhood~\cite{gat, gatv2, aero}. Thus, when new layers are added to GCN, further neighboring nodes are aggregated as a weighted sum of their representations. Therefore, the standard graph attention is prone to over-smoothing as the number of layers increases. Although hop attention could help diminish the effect of over-smoothed features at deeper layers, it cannot select information from specific edges in a hop. 

This study proposes a channel-wise attentive message-passing mechanism, coined CHAT-GNN, that can adaptively learn which information to send from a neighbor to the source node based on different hops, edges, and feature channels. The performance of the proposed CHAT-GNN is compared with various baselines and benchmark graph datasets. We present a GNN architecture that utilizes our channel-attentive message-passing mechanism, and we experimentally show that it outperforms the strong baselines in the literature. In addition, we show that CHAT-GNN mitigates over-smoothing at deeper layers while most of the baselines severely suffer from it. Finally, we evaluate the performance of CHAT-GNN in different settings and analyze the effect of additional components through an ablation study. The contributions of this study are outlined below:
\begin{itemize}
    \item We approach graph attention from a channel-wise perspective and propose a message-passing mechanism that applies channel-wise updates. 
    \item The proposed channel-wise attention can be integrated into any message-passing GNN architecture trained for any task. 
  \item Extensive analyses are conducted, providing empirical evidence that the proposed CHAT-GNN improves the node classification performance and alleviates over-smoothing.
 \item The visualization of the attention weights shows that the channel-wise attention of CHAT-GNN can, in fact, adapt to different neighbors and hops.
\end{itemize}
Experiments are conducted with various benchmark datasets commonly used in the GNN literature. The performance of CHAT-GNN is compared with commonly reported and recent baselines. Resistance of CHAT-GNN to over-smoothing is investigated and compared with several baselines. 

\section{Related Work}
The related work is divided into studies with graph attention and gating, studies dealing with over-smoothing, and studies working with heterophilous graphs, which are severely impacted by over-smoothing.

\subsection{Attention \& Gating}
Attention-based graph neural networks~\cite{gat, gatv2, aero, gtrans, fagcn} have a growing popularity in various graph-based tasks. Graph attention applies the well-known attention mechanism~\cite{luong, bahdanau, transformer} over the neighborhood of a node to obtain an edge-specific weighting that enables more expressive message aggregation instead of the uniform aggregation in standard GCN. Moreover, some studies propose attending over the message-passing layers so that each layer gets a different weight~\cite{gprgnn, dagnn, aero}. These two types of graph attention are called edge attention and hop attention~\cite{aero}. CHAT-GNN extends graph attention by learning a channel-wise weighting mechanism that determines the rate of the information flow for each channel in an edge. 

The proposed framework is also closely related to the gating mechanism in neural networks. The recurrent neural networks utilize the gating mechanism to control the information flow between consecutive time steps~\cite{gru, lstm}. FAGCN~\cite{fagcn} employs a self-gating mechanism that adjusts the frequency of the underlying filtering in message-passing. On the other hand, Gradient gating~\cite{g2gnn} uses graph gradients to control the information flow in a multi-rate fashion. Ordered GNN~\cite{gonn} aims to order the information coming from different hops by considering the message-passing scheme for a node as a rooted tree growing with the hops. The model uses a gating mechanism to separate the information from different hops in the message vector. Thus, the primary motivation is to prevent the mixing of node features from different hops~\cite{gonn}. 

The CHAT-GNN has similarities and differences between the standard attention and gating mechanisms. The proposed channel-wise attention aims to control high and low-frequency changes in the feature channels while learning to attend to each neighboring node, layer, and feature channel differently. Thus, the proposed message-passing framework extends the graph attention and gate ideas to each direction in the GNN's latent space.

\subsection{Over-smoothing}
Over-smoothing is an issue in GNNs that results in indistinguishable node representations when stacking many message-passing layers~\cite{oversmoothing, gnn_exp, measure_smooth}. This issue stems from the message-passing layers acting as low-pass filters and over-smoothing the feature vectors of closer nodes~\cite{oversmoothing, measure_smooth, lowpass}. As a result, adding layers to GNNs with such message-passing mechanisms induces underfitting. However, graphs can have arbitrary topological structures, and the task of interest may require leveraging the underlying deep and complex connections in the graph. Many studies aim to mitigate the over-smoothing issue from various perspectives, such as data augmentation and graph topology optimization~\cite{dropedge, measure_smooth}. For instance, Rong {\it et al.} proposed randomly removing some edges during training to prevent overfitting and over-smoothing~\cite{dropedge}. On the other hand, Chen {\it et al.} propose topological metrics to quantify over-smoothing and a framework that optimizes the proposed metrics~\cite{measure_smooth}. 

More recent studies focus on the connection between over-smoothing and the information flow during message passing, and thus, they address over-smoothing with gating~\cite{fagcn, g2gnn, gonn}, and attention~\cite{aero, dagnn, gprgnn}. The analyses in this study show that the performance of the proposed CHAT-GNN framework is less vulnerable than the baselines against deepening the GNN. 

\subsection{Learning on Heterophilous Graphs}
Node classification on heterophilous graphs is widely studied~\cite{h2gcn, gprgnn, gonn, platonov}. A graph is considered heterophilous if most of its edges are between the nodes from different classes. Learning on heterophilous graphs is challenging since the standard message-passing mechanisms used in GCN, GAT, and similar models enforce the representation of adjacent nodes to be close to each other. Consequently, the discriminability of the node representations is sensitive to message-passing operations in heterophilous graphs.

Some studies on heterophilous graphs focus on handling neighborhood orders differently by learning hop weights~\cite{gprgnn, aero} and separating the representation of different hops~\cite{h2gcn}. Separating the node's representation and its neighborhood representation during the combine phase of message-passing is also emphasized by some studies~\cite{h2gcn, platonov}. The separation of representations prevents us from mixing the two representations that are supposed to be dissimilar in heterophilous graphs~\cite{h2gcn}. The GraphSAGE~\cite{sage} model also has a separate embedding mechanism for the node and its neighborhood. Platonov {\it et al.} showed that GraphSAGE~\cite{sage} performs better than GCN and GAT in heterophilous graph datasets proposed by the authors~\cite{platonov}. We also follow this principle by applying separate learnable affine transformations to the node embeddings and their neighborhood representation. Similarly to prior work, we also observe that allowing this mixing does not hurt the performance in graphs with high homophily.
\section{Method}

\begin{figure*}[t!]
    \centering
\includegraphics[width=1\linewidth]{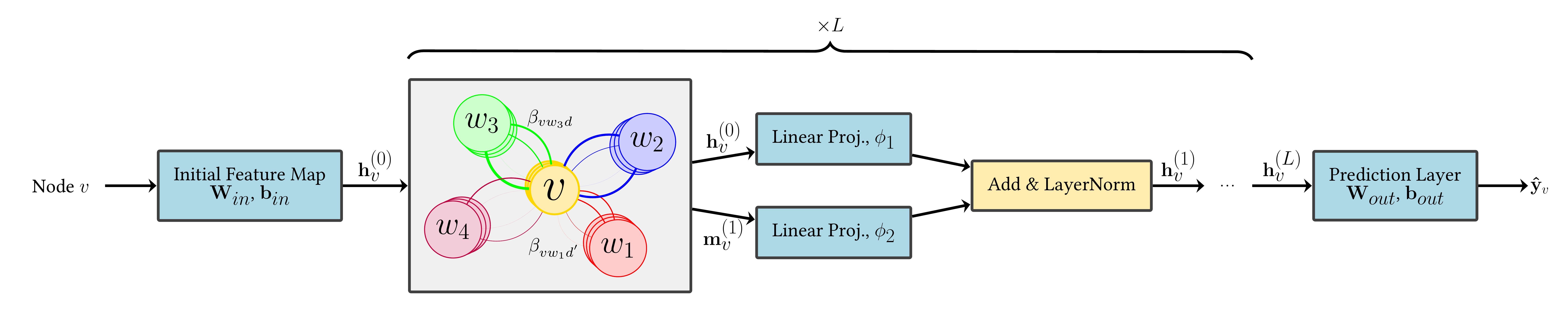}
    \caption{The architecture of CHAT-GNN. After a shared input layer projects the node features,  $L$ channel-attentive message-passing layers are included. Each message-passing operation is followed by linear projections and layer normalization. The final prediction is obtained by feeding the output of the $L$-th layer to an output layer.}
    \label{fig:model_architecture}
\end{figure*}

In this section, we provide the details of CHAT-GNN. We start with some basic definitions related to graphs and GNNs. We are interested in undirected graphs without self-loops and edge weights. Such graphs can be denoted as $\mathcal{G} = (\mathcal{V}, \mathcal{E})$, where $\mathcal{V}$ is the set of nodes and $\mathcal{E}$ is the set of edges. Let $N$ be the number of nodes in the graph. Alternatively, we can also represent a graph by the adjacency matrix $\mathbf{A} \in \mathbb{R}^{N \times N}$. Each node has an input feature vector $\mathbf{x} \in \mathbb{R}^F$, and all of them together form an input feature matrix $\mathbf{X} \in \mathbb{R}^{N \times F}$, where $F$ is the dimensionality of the node feature. Since the node classification task is considered, each node must have a label $\mathbf{y} \in \{ 0, 1 \}^K$, where $K$ is the number of available classes in the dataset. We also denote the neighbor set of a node $v \in \mathcal{V}$ by $\mathcal{N}(v)$.


Most GNNs utilize the message-passing paradigm due to its simple local operations to calculate representations on graphs efficiently~\cite{gilmer, battaglia}. The basic idea of message-passing is calculating node representations by receiving "messages" from immediate neighbors, aggregating them into a single representation of the whole neighborhood, and repeating this process as many times as required to allow information to flow from distant nodes. Assume that $\mathbf{h}_v^{(k)} \in \mathbb{R}^{D}$ is the representation of the node $v$ at layer $k$.  The message-passing mechanism in GNNs for the layer $k+1$ can be formalized as follows:
\begin{equation}
    \mathbf{m}_{vw}^{(k+1)} = \text{MSG}^{(k+1)}(\mathbf{h}_v^{(k)}, \mathbf{h}_w^{(k)})
\end{equation}
\begin{equation}
    \mathbf{m}_v^{(k+1)} = \text{AGG}^{(k+1)} \left(  \left\{ \mathbf{m}_{vw}^{(k+1)} \ | \ w \in \mathcal{N}(v) \right\} \right )
    \label{eq:mp}
\end{equation}
\begin{equation}
    \mathbf{h}_v^{(k+1)} = \text{CMB}^{(k+1)}\left( \mathbf{h}_v^{(k)}, \mathbf{m}_v^{(k+1)} \right)
\end{equation}
where $\text{AGG}^{(k+1)}$ is any permutation-invariant aggregation function, $\text{MSG}^{(k+1)}$ is the message function calculating the message that will passed from node $w$ to node $v$, and finally, $\text{CMB}^{(k+1)}$ combines the node's current representation and the neighborhood representation, $\mathbf{m}_v^{(k+1)}$, into a single representation as the updated representation of node $v$ at layer $k+1$. The design choices for these three fundamental functions distinguish the message-passing GNNs in the literature~\cite{gcn, gat, gatv2, sage, fagcn}. 

In this study, we are specifically interested in the $\text{MSG}(\cdot, \cdot)$ function since it has been shown that designing message functions that can learn to output specialized messages between different nodes helps in various tasks~\cite{gat, gatv2, fagcn, aero}. Attention-based GNNs~\cite{gat, gatv2, aero, fagcn} use message functions that learn a weighting between pairs of nodes as follows:
\begin{equation}
    \text{MSG}(\mathbf{h}_v, \mathbf{h}_w) =  \alpha(\mathbf{h}_v, \mathbf{h}_w) \mathbf{h}_w
\end{equation}
where $\alpha: \mathbb{R}^D \times \mathbb{R}^D \to \mathbb{R}$ is the weighting function that outputs a scalar, called attention coefficients or attention scores indicating the importance of the neighbor $w$ to the node $v$. The function $\alpha(\cdot, \cdot)$ is often designed using fully connected layers~\cite{gat,  gatv2, aero, fagcn}. Note that we assume unnormalized attention functions to indicate that the output depends on $\mathbf{h}_v$ and $\mathbf{h}_w$. These attention weights can be normalized over the node's neighborhood using an activation function, such as softmax, so that the sum of the scores always equals one~\cite{gat, gatv2}. The following section explains the proposed channel-wise attention for the neighboring nodes.

\subsection{Channel-Attentive Message-Passing}
\label{sub:chat}
Although traditional attention-based message-passing performs well on some benchmarks, it still suffers from over-smoothing and fails on many heterophilous graphs. Let $\alpha_{vw} = \alpha(\mathbf{h}_v, \mathbf{h}_w)$ be the attention score between a pair of adjacent nodes $v$ and $w$. The fact that each feature channel gets multiplied by the same attention score, $\alpha_{vw}$, and the distinction between messages passed from $v$ to the arbitrary neighbors $w$ and $u$ being simply defined with the ratio $\alpha_{vw} / \alpha_{vu}$ prevent the model from learning distinctive representations of each neighbor during aggregation. In addition, we may want to diminish the contribution of a feature from a specific neighbor and increase the effect of that feature from another neighbor. Further, we may want to control the rate of information flow from each feature channel just like the traditional attention-based models controlling the contribution of each neighbor. With this motivation, we consider the weighting functions of the form $\beta: \mathbb{R}^D \times \mathbb{R}^D \to \mathbb{R}^D$. We denote $\beta_{vw} = \beta(\mathbf{h}_v, \mathbf{h}_w)$ as the attention weight vector between the pair of nodes $v$ and $w$, and $\beta_{vwd}$ denotes the importance of the feature channel $d$ for the specific node pair $v$ and $w$. Accordingly, we reformulate our message function to use the channel-wise importance as follows:
\begin{equation}
    \text{MSG}(\mathbf{h}_v, \mathbf{h}_w) =  \beta(\mathbf{h}_v, \mathbf{h}_w) \odot \mathbf{h}_w
\end{equation}
where $\odot$ denotes the element-wise multiplication, also known as the Hadamard product.

The channel-attentive function $\beta$ of CHAT-GNN is designed as a fully connected module optimized during training in an end-to-end fashion. The $\beta_{vwd}$ values are crucial for the expressivity of the message-passing phase. Unlike the standard attention with output values between zero and one, we allow negative values to capture negative interactions between the neighboring nodes that help mitigate over-smoothing~\cite{fagcn, unifying}. Therefore, the proposed channel-attentive function is computed as follows:
\begin{equation}
    \beta(\mathbf{h}_v, \mathbf{h}_w) = \text{tanh} \left( \mathbf{W}_1 \mathbf{h}_v + \mathbf{W}_2 \mathbf{h}_w \right)
\end{equation}
where $\mathbf{W}_1, \mathbf{W}_2 \in \mathbb{R}^{D \times D}$ are trainable weight matrices multiplied by the source and the neighbor nodes, respectively. The contribution of allowing negative values can be explained by an analogy with the filtering in image processing. A low-pass filter applied to local regions on an image has a smoothing effect that reduces the noise in the image. While a high-pass filter emphasizes high-frequency regions, such as edges, resulting in a sharpening effect. Similarly, we employ the message-passing mechanism that acts on local regions of the graph, which are neighborhoods of nodes. Both filtering approaches could be desirable in an arbitrary graph structure, especially in large and complex real-world graphs. The proposed message-passing formulation resembles a hybrid low and high-pass filtering applied on the same region but on different color channels.

The full expression of the aggregation part of the channel-attentive message-passing mechanism can be formalized as follows:
\begin{equation}
    \mathbf{m}_v = \sum_{w \in \mathcal{N}(v)} \frac{1}{\sqrt{d_v d_w}} \beta(\mathbf{h}_v, \mathbf{h}_w) \odot \mathbf{h}_w
    \label{eq:chat}
\end{equation}
where $\mathbf{m}_v$ is the neighborhood representation obtained by the channel-attentive message-passing mechanism. Note that we use the summation as the aggregation operator $\text{AGG}$ in Equation~\eqref{eq:mp} and the inverse square root of degrees for scaling~\cite{gcn, fagcn}. The equation above shows that each feature channel is attended differently, and the attention weights are computed considering the neighboring features. Thus, the network can adjust the channel-wise message-passing mechanism by informing each feature channel about other feature channels. 


\subsection{The Combine Phase}

We consider the following expression for the combine operation, $\text{CMB}^{(k+1)}$, in message-passing for an arbitrary layer $k+1$:
\begin{equation}
    \mathbf{\tilde{h}}_v^{(k+1)} = \phi_1^{(k+1)}(\mathbf{h}_v^{(k)}) + \phi_2^{(k+1)}(\mathbf{m}_v^{(k+1)})
\end{equation}
where $\phi_1: \mathbb{R}^D \to \mathbb{R}^D$ and $\phi_2: \mathbb{R}^D \to \mathbb{R}^D$ are linear transformations applied on the current node representation and the neighborhood representation, respectively. As hinted in the literature~\cite{h2gcn, sage, platonov}, separately projecting the node and neighborhood representations performs better than directly mixing the two representations. Furthermore, Zhu {\it et al.}~\cite{h2gcn} showed that, in some cases, this approach results in a more capable model for heterophilous settings. In our experiments, we observe that adding separate transformations, $\phi_1, \phi_2$, improves the performance in some graphs in both high and low homophily settings. However, we achieve the best results for most datasets without separate linear projection layers. Therefore, we treat this choice as a hyperparameter to tune between the identity function $\phi_1 (\mathbf{x}) = \mathbf{x}$ and a single linear layer $\phi_1(\mathbf{x}) = \mathbf{W}_1^{'} \mathbf{x}$ discussed with an ablation study in Section~\ref{sub:ablation}.

\subsection{CHAT-GNN: The Whole Architecture}
\label{sec:whole}
Now, we present the proposed CHAT-GNN architecture illustrated in Figure~\ref{fig:model_architecture}. Let us consider the forward propagation for an arbitrary node $v$. Assume that the input features of node $v$ are denoted as $\mathbf{x}_v$. First, the input node features are projected into a hidden representation with a non-linear layer as follows:
\begin{equation}
    \mathbf{h}_v^{(0)} =\text{ReLU}\left( \mathbf{W}_{\text{in}} \mathbf{x}_v + \mathbf{b}_{\text{in}} \right)
\end{equation}
where $\mathbf{W}_{\text{in}} \in \mathbb{R}^{D \times F}$, $\mathbf{b}_{\text{in}} \in \mathbb{R}^{D \times 1}$ are weight and bias of the initial feature map layer. We denote $\mathbf{h}_v^{(0)}$ as the initial input to the message-passing layers. The initial feature projection is beneficial to reducing the large number of sparse input features and learning fine-grained features that will be passed to the message-passing mechanism. Following the footsteps of previous studies~\cite{fagcn, gcn2, platonov}, we observed that applying residual connection between the initial features and the message-passing layers helps speed up the training process since the model may want to access the initial fine-grained features in deeper message-passing layers. Following the initial residual connection, we perform layer normalization to avoid exploding hidden representations:
\begin{equation}
    \mathbf{h}_v^{(k+1)} = \text{LayerNorm}^{(k+1)} \left(\mathbf{\tilde{h}}_v^{(k+1)} + \mathbf{h}_v^{(0)}\right)
\end{equation}
where $\text{LayerNorm}(\cdot)^{(k+1)}$ is the layer normalization at the ($k+1$)-th message-passing layer, applied on the feature channels~\cite{layernorm}. Stacking many message-passing layers increases the norm of node representations that cause instabilities and overfitting. Therefore, GNNs benefit from various normalization techniques that help tackle over-smoothing~\cite{pairnorm, contranorm}. For instance, AERO-GNN uses a predetermined series of scalars to scale the output representations~\cite{aero}. On the other hand, Platonov {\it et al.}~\cite{platonov} used layer normalization and residual connections to boost the performance of GCN, GAT, and GraphSAGE on heterophilous graphs. We also observe the benefit of layer normalization, especially in deeper architectures, for large enough heterophilous graphs. For extremely small graphs and graphs with high homophily, the overfitting problem occurs since the neighborhood patterns can easily be memorized.

Finally, an output layer is applied to the final node representations to obtain the predictions:
\begin{equation}
    \mathbf{z}_v = \mathbf{W}_{\text{out}} \mathbf{h}_v^{(L)} + \mathbf{b}_{\text{out}}
\end{equation}
where $\mathbf{W}_{\text{out}} \in \mathbb{R}^{K \times D}$, $\mathbf{b}_{\text{out}} \in \mathbb{R}^{K \times 1}$ are the weight and bias of the output layer and $L$ is the number message-passing layers in CHAT-GNN. The proposed architecture is trained for the node classification task. Therefore, the softmax function is used as the output layer's activation. The cross-entropy loss function below is used to optimize the proposed and baseline models:
\begin{equation}
    L = -\frac{1}{N_{\text{train}}} \sum_{n=1}^{N_{\text{train}}}\sum_{k=1}^{K} y_{nk} \cdot \text{log}(\hat{y}_{nk})
\end{equation}
where $N_{\text{train}}$ denotes the number of training nodes, $K$ denotes the number of classes, and $\hat{y}_{nk}$ and $y_{nk}$ are the predicted and true labels of node $n$, respectively.

\subsection{Comparison with Multi-Head Attention}

As discussed in previous sections, our approach is essentially an extension of standard attention to each feature channel. Multi-head attention might seem to lie between these two ideas. Multi-head attention maps the input representations to different subspaces to learn more diverse output representations~\cite{transformer, gat}. The learned representation in each head is concatenated to generate the final output. However, it still applies the standard attention mechanism in each head. For example, multi-head graph attention  with $K$ heads is illustrated as follows~\cite{gat}:
\begin{equation}
    \mathbf{m}_{v} = \concat_{k=1}^K \sigma \big (  \sum_{w \in \mathcal{N}(v)} \alpha_{vwk} \mathbf{W}_k \mathbf{h}_{w}  \big )
\end{equation}
where $\alpha_{vwk}$ is the attention score between nodes $v$ and $w$ for attention head $k$, and $\mathbf{W}_k$ is the weight matrix that extracts features for head $k$ to attend on.

On the other hand, in channel-wise attention, as formalized in~\eqref{eq:chat}, we calculate channel-wise importance by considering the original node representations, and all feature channels are able to interact with each other. Also, the learned attention scores are used to weigh the input representations. Therefore, channel-wise attention differs from other attention-based mechanisms, offering a more flexible information flow during message-passing.

\subsection{Theoretical Analysis}
In this section, we discuss the effectiveness of channel-attentive message-passing by providing some theoretical findings. We start with the notion of local smoothness. \textit{Local variation} of node $v$ according to graph signal $\mathbf{X}$ is defined as follows \cite{emerging_gsp}:
\begin{equation}
    \mathcal{E}_v(\mathbf{X}) = \sum_{w \in \mathcal{N}(v)} \| \mathbf{x}_v - \mathbf{x}_w \|^2
\end{equation}
Then, the \textit{Dirichlet Energy} is defined as the sum of local variations across all nodes, which can be seen as a measure of global smoothness \cite{emerging_gsp, graphcon}:
\begin{equation}
    \mathcal{E}(\mathbf{X}) = \frac{1}{N} \sum_{v \in \mathcal{V}} \mathcal{E}_v(\mathbf{X})
    \label{eq:dirichlet}
\end{equation}
Thus, the global smoothness of the node features depends on the local variations. The following proposition indicates that the change in the local variation between consecutive GNN layers is bounded by the distance between the aggregated messages.

\begin{proposition}
    Let us assume that we have a GNN in which the (k+1)-th layer's update is as follows: $\mathbf{h}_v^{(k+1)} = \mathbf{h}_v^{(k)} + \mathbf{m}_v^{(k+1)}$, where $\mathbf{m}_v^{(k+1)}$ is the output of a message-passing layer. Then, the change in the local variation of node $v$, $\Delta_{(k+1)} \mathcal{E}_v = \mathcal{E}_v^{(k+1)} - \mathcal{E}_v^{(k)}$ satisfies the following inequality:
    \begin{equation}
        \Delta_{(k+1)} \mathcal{E}_v \leq \sum_{w \in \mathcal{N}(v)} (\delta_{wv}^{(k+1)})^2 + 2c\delta_{wv}^{(k+1)}
    \end{equation}
    where $\delta_{wv}^{(k+1)} = \| \mathbf{m}_w^{(k+1)} - \mathbf{m}_v^{(k+1)} \|$ and $c = \max \| \mathbf{h}_w^{(k)} - \mathbf{h}_v^{(k)} \|$.

\label{prop1}
\end{proposition}
\begin{IEEEproof}
\begin{IEEEeqnarray}{lCl}
\Delta_{(k+1)} \mathcal{E}_v &=& \sum_{w \in \mathcal{N}(v)} 
\Big ( \| \mathbf{h}_w^{(k)} + \mathbf{m}_w^{(k+1)} - 
(\mathbf{h}_v^{(k)} + \mathbf{m}_v^{(k+1)}) \|^2\IEEEnonumber\\
&&- \| \mathbf{h}_w^{(k)} - \mathbf{h}_v^{(k)} \|^2 \Big )\IEEEnonumber\\
&=& \sum_{w \in \mathcal{N}(v)} \Big [ \Big ( 
    \| \mathbf{h}_w^{(k)} + \mathbf{m}_w^{(k+1)} - 
(\mathbf{h}_v^{(k)} + \mathbf{m}_v^{(k+1)}) \| \IEEEnonumber\\
&&- \| \mathbf{h}_w^{(k)} - \mathbf{h}_v^{(k)} \|
\Big )\Big ( 
    \| \mathbf{h}_w^{(k)} + \mathbf{m}_w^{(k+1)} - \IEEEnonumber\\
&&(\mathbf{h}_v^{(k)} + \mathbf{m}_v^{(k+1)}) \|  + \| \mathbf{h}_w^{(k)} - \mathbf{h}_v^{(k)} \|
\Big )   \Big ]
\end{IEEEeqnarray}

Using the triangle inequality, we get:
\begin{IEEEeqnarray}{lCl}
\Delta_{(k+1)} \mathcal{E}_v &\leq& \sum_{w \in \mathcal{N}(v)} 
\Big [ \Big ( 
    \| \mathbf{m}_w^{(k+1)} - \mathbf{m}_v^{(k+1)} \|
\Big )\IEEEnonumber\\
&&\Big ( 
    \| \mathbf{m}_w^{(k+1)} - \mathbf{m}_v^{(k+1)} \| 
    + 2\| \mathbf{h}_w^{(k)} - \mathbf{h}_v^{(k)} \|
\Big )   \Big ]\IEEEnonumber\\
&=& \sum_{w \in \mathcal{N}(v)} (\delta_{wv}^{(k+1)})^2 + 2c\delta_{wv}^{(k+1)}
\end{IEEEeqnarray}
\end{IEEEproof}

Proposition~\ref{prop1} shows that the upper bound on the change in the local variation is determined by how different the message vectors of neighboring nodes are. The over-smoothing effect begins when a GNN layer's output for a node is similar to its neighbors. Next, we present another proposition that compares the standard graph attention and our proposed channel-wise graph attention in terms of how likely the $\delta_{wv} = \| \mathbf{m}_w - \mathbf{m}_v \|$  would be non-zero.

\begin{proposition}
    Assume that we have a pair of nodes $v$ and $w$ that satisfy $\mathcal{N}(v) = \mathcal{N}(w)$, and positive node features $\mathbf{X}$. Further, assume that message vectors with standard and channel-wise attention are calculated as $\mathbf{m}_v =  \sum_{k \in \mathcal{N}(v)} \alpha_{vk} \mathbf{x}_k $ and $\mathbf{m}_v' = \sum_{k \in \mathcal{N}(v)} \beta_{vk} \odot \mathbf{x}_k  $, respectively. Lastly, we assume that each unnormalized attention score $\alpha_{vk}$ and $\beta_{vkd}$ are all independent random variables. Then, we have the following inequalities satisfied:
    \begin{equation}
        \Pr(\delta_{m_{vw}}^* > 0 ) \leq | \mathcal{N}(v) | P^*
    \end{equation}
    \begin{equation}
        \Pr(\delta_{m_{vw}'}^* > 0 ) \leq | \mathcal{N}(v) | D P^*
    \end{equation}
    where $\delta_{m_{vw}}^* = \sum_{k \in \mathcal{N}(v)} |\alpha_{vk} - \alpha_{wk}| \| \mathbf{x}_k \|$ and $\delta_{m_{vw}'}^* = \sum_{k \in \mathcal{N}(v)} \| (\beta_{vk} - \beta_{wk}) \odot \mathbf{x}_k \|$ are the respective upper bounds of $\delta_{vw}$ under the standard and channel-wise attention settings, $D$ is the dimensionality of feature vectors and channel-wise attention vector $\beta$, and $P^*$ is the maximum possible value that $\Pr( | \alpha - \alpha' | > 0 )$ can attain for two arbitrary attention scores $\alpha$ and $\alpha'$. 

\label{prop2}
\end{proposition}
\begin{IEEEproof}
    We begin with the first inequality. We can see that $\delta_{m_{vw}}^*$ is greater than zero if for at least one neighbor $k$, we have $|\alpha_{vk} - \alpha_{wk}| \neq 0$. Therefore, using Boole's inequality, we have the following:
    \begin{IEEEeqnarray}{lCl}
        \Pr(\delta_{m_{vw}}^* > 0 ) &=& \Pr( \bigcup_{k \in \mathcal{N}(v)} \{  |\alpha_{vk} - \alpha_{wk}| \neq 0 \} ) \IEEEnonumber\\
        &\leq& \sum_{k \in \mathcal{N}(v)} \Pr(|\alpha_{vk} - \alpha_{wk}| \neq 0)\IEEEnonumber\\
        &\leq& | \mathcal{N}(v) | P^*
    \end{IEEEeqnarray}

    Now, we prove the second equality. Similarly, we see that the $\delta_{m_{vw}'}^*$ is greater than zero if for at least one dimension $d$ for one neighbor $k$, we have $|\beta_{vkd} - \beta_{wkd}| \neq 0$. Again, we utilize Boole's inequality as follows:
    \begin{IEEEeqnarray}{lCl}
        \Pr(\delta_{m_{vw}'}^* > 0 ) &=& \Pr( \bigcup_{k \in \mathcal{N}(v)} \{  \| \beta_{vk} - \beta_{wk} \| \neq 0 \} ) \IEEEnonumber\\
        &\leq& \sum_{k \in \mathcal{N}(v)} \Pr(\| \beta_{vk} - \beta_{wk} \| \neq 0 )\IEEEnonumber\\
        &=& \sum_{k \in \mathcal{N}(v)} \Pr( \bigcup_{d=1}^D \{  | \beta_{vkd} - \beta_{wkd} | \neq 0 \} ) \IEEEnonumber\\
        &\leq& \sum_{k \in \mathcal{N}(v)} \sum_{d=1}^D \Pr(   | \beta_{vkd} - \beta_{wkd} | \neq 0  )\IEEEnonumber\\
        &\leq& | \mathcal{N}(v) | D P^*
    \end{IEEEeqnarray}
\end{IEEEproof}

Therefore, in the worst case where two nodes $v$ and $w$ have exactly the same neighborhood, the channel-attentive attention mechanism is more likely to avoid collapsed message vectors. The following sections present the experimental setup and results on well-known benchmark datasets. 
\section{Experiments}
This section reports the node classification results of experiments on 14 benchmark datasets. The performance of CHAT-GNN is compared with various well-known baselines, including recent studies tackling the over-smoothing problem. 

\begin{table}[t]
\centering
\caption{Statistics of the benchmark datasets. Homophily metric proposed by Lim et al.~\cite{lim} is provided.}
\begin{tabular}{lllll}
\toprule
\textbf{dataset} & \textbf{\# nodes} & \textbf{\# edges} & \textbf{\# features} & \textbf{homophily} \\
\midrule
$\text{roman-empire}$ & $\text{22,662}$ & $\text{32,927}$ & $\text{300}$ & $\text{0.02}$ \\
$\text{amazon-ratings}$ & $\text{24,492}$ & $\text{93,050}$ & $\text{300}$ & $\text{0.13}$ \\
$\text{minesweeper}$ & $\text{10,000}$ & $\text{39,402}$ & $\text{7}$ & $\text{0.01}$ \\
$\text{tolokers}$ & $\text{11,758}$ & $\text{519,000}$ & $\text{10}$ & $\text{0.17}$ \\
$\text{questions}$ & $\text{48,921}$ & $\text{153,540}$ & $\text{301}$ & $\text{0.09}$ \\
$\text{chameleon}$ & $\text{2,277}$ & $\text{36,051}$ & $\text{2,325}$ & $\text{0.06}$ \\
$\text{squirrel}$ & $\text{5,201}$ & $\text{216,933}$ & $\text{2,089}$ & $\text{0.04}$ \\
$\text{cornell}$ & $\text{183}$ & $\text{295}$ & $\text{1,703}$ & $\text{0.13}$ \\
$\text{actor}$ & $\text{7,600}$ & $\text{29,926}$ & $\text{932}$ & $\text{0.00}$ \\
$\text{texas}$ & $\text{183}$ & $\text{309}$ & $\text{1,703}$ & $\text{0.00}$ \\
$\text{wisconsin}$ & $\text{251}$ & $\text{499}$ & $\text{1,703}$ & $\text{0.08}$ \\
$\text{cora}$ & $\text{2,708}$ & $\text{10,556}$ & $\text{1,433}$ & $\text{0.77}$ \\
$\text{citeseer}$ & $\text{3,327}$ & $\text{9,104}$ & $\text{3,703}$ & $\text{0.63}$ \\
$\text{pubmed}$ & $\text{19,717}$ & $\text{88,648}$ & $\text{500}$ & $\text{0.66}$ \\
\bottomrule
\end{tabular}
\label{tab:dataset_stats}
\end{table}

\subsection{Datasets}
We train CHAT-GNN with commonly used benchmark datasets for the node classification task. Among the benchmark datasets, three citation network datasets with high homophily, Cora, Citeseer, and Pubmed~\cite{ccp}, are considered. We use the full split versions of these datasets in which the nodes not in validation and test sets are used for training~\cite{fastgcn}. Chameleon, Squirrel, and Actor, heterophilous Wikipedia networks, are also included~\cite{geom}. We also consider Cornell, Texas, and Wisconsin, small-scale web networks with high heterophily~\cite{geom}. Finally, we use a recent set of extremely heterophilous benchmark datasets, namely roman-empire, amazon-ratings, minesweeper, tolokers, and questions designed to evaluate graph representation learning in heterophilous settings~\cite{platonov}. The dataset statistics are provided in Table~\ref{tab:dataset_stats} with the homophily metric proposed by Lim {\it et al.}~\cite{lim}.

We convert all graphs to undirected graphs by adding edges in both directions for overall comparison. Feature normalization that helps stabilize the training process is applied for datasets with bag-of-words features~\cite{ccp, geom}. The node features provided in the datasets by Platonov {\it et al.}~\cite{platonov} are used without any modification. 
\subsection{Baselines}
We have a diverse set of baselines from the GNN literature. GCN~\cite{gcn}, GAT~\cite{gat}, and GATv2~\cite{gatv2} are included as the standard models in the literature. GCN-II~\cite{gcn2} and Jumping Knowledge Network (JKNet)~\cite{jknet} are early attempts at deep learning on graphs. H2GCN~\cite{h2gcn} is an improvement in learning on graphs with high heterophily. FAGCN~\cite{fagcn} tries to overcome over-smoothing with high-pass filters. AERO-GNN~\cite{aero} and Ordered GNN~\cite{gonn} are recent models that can scale to many layers and mitigate over-smoothing. We also include a Multilayer Perceptron (MLP), which only uses the node features, to have an idea of how much it can learn without using the graph structure.

\begin{table*}[t!]
\centering
\caption{Node classification performance on benchmark datasets. All the models are trained from scratch. Official implementation of the baselines and hyperparameter setup are used when available. }
\resizebox{\linewidth}{!}{
\begin{tabular}{ccccccccccccccc}
\toprule
 & roman-emp. & amazon-rat. & minesweeper & tolokers & questions & chameleon & squirrel & cornell & actor & texas & wisconsin & cora & citeseer & pubmed \\
\midrule
MLP & $67.0 \pm 0.6$ & $45.2 \pm 0.5$ & $51.0 \pm 1.3$ & $71.9 \pm 0.7$ & $70.5 \pm 1.0$ & $49.6 \pm 2.3$ & $33.2 \pm 1.1$ & $73.3 \pm 7.1$ & $35.6 \pm 1.5$ & $78.9 \pm 3.8$ & $85.3 \pm 3.2$ & $76.9 \pm 0.0$ & $75.4 \pm 0.0$ & $89.1 \pm 0.0$ \\
GCN & $51.9 \pm 0.5$ & $48.4 \pm 0.5$ & $72.3 \pm 1.0$ & $75.2 \pm 1.3$ & $76.3 \pm 1.0$ & $64.7 \pm 2.1$ & $47.7 \pm 1.9$ & $50.0 \pm 9.2$ & $30.6 \pm 0.5$ & $63.8 \pm 4.2$ & $62.2 \pm 5.1$ & $87.3 \pm 0.3$ & $78.7 \pm 0.4$ & $88.7 \pm 0.2$ \\
GAT & $60.0 \pm 0.9$ & $45.9 \pm 0.7$ & $86.3 \pm 0.9$ & $74.4 \pm 0.8$ & $70.9 \pm 1.2$ & $68.1 \pm 1.7$ & $52.0 \pm 1.4$ & $45.9 \pm 7.1$ & $29.8 \pm 1.0$ & $59.7 \pm 6.4$ & $64.9 \pm 5.2$ & $\underline{87.6 \pm 0.4}$ & $78.6 \pm 0.4$ & $87.0 \pm 0.3$ \\
GATv2 & $52.1 \pm 1.9$ & $44.2 \pm 0.6$ & $73.4 \pm 1.0$ & $73.4 \pm 0.9$ & $73.3 \pm 1.0$ & $63.9 \pm 1.5$ & $47.2 \pm 1.5$ & $46.2 \pm 7.3$ & $29.2 \pm 1.0$ & $61.4 \pm 5.9$ & $61.2 \pm 4.3$ & $\underline{87.6 \pm 0.4}$ & $78.5 \pm 0.3$ & $86.7 \pm 0.3$ \\
GCN-II & $82.5 \pm 0.7$ & $52.2 \pm 0.5$ & $90.5 \pm 0.6$ & $77.5 \pm 1.0$ & $74.6 \pm 0.9$ & $66.1 \pm 2.0$ & $49.4 \pm 1.4$ & $68.7 \pm 7.8$ & $\underline{36.0 \pm 0.9}$ & $76.8 \pm 6.9$ & $83.5 \pm 5.2$ & $86.7 \pm 0.5$ & $78.4 \pm 0.5$ & $\mathbf{90.6 \pm 0.5}$ \\
JKNet & $79.2 \pm 0.6$ & $46.5 \pm 0.5$ & $88.2 \pm 0.7$ & $80.7 \pm 1.3$ & $77.5 \pm 0.9$ & $66.1 \pm 1.6$ & $54.0 \pm 1.5$ & $73.5 \pm 3.1$ & $35.3 \pm 0.6$ & $79.2 \pm 5.0$ & $83.1 \pm 4.5$ & $85.9 \pm 0.5$ & $77.8 \pm 0.7$ & $89.4 \pm 0.3$ \\
H2GCN & $79.5 \pm 0.6$ & $50.2 \pm 0.4$ & $90.2 \pm 0.5$ & OOM & $71.3 \pm 1.1$ & $58.0 \pm 1.6$ & $33.4 \pm 1.9$ & $71.9 \pm 5.6$ & $35.5 \pm 1.1$ & $82.2 \pm 5.2$ & $\underline{86.3 \pm 3.6}$ & $87.2 \pm 0.3$ & $78.0 \pm 0.4$ & $90.1 \pm 0.1$ \\
FAGCN & $71.5 \pm 0.5$ & $48.4 \pm 0.7$ & $87.8 \pm 0.9$ & $77.4 \pm 1.2$ & $75.9 \pm 1.6$ & $61.0 \pm 1.5$ & $38.6 \pm 2.5$ & $\underline{76.2 \pm 2.6}$ & $35.3 \pm 0.9$ & $83.3 \pm 6.4$ & $81.4 \pm 4.2$ & $\mathbf{88.4 \pm 0.3}$ & $78.3 \pm 0.5$ & $90.3 \pm 0.2$ \\
AERO-GNN & $80.5 \pm 0.9$ & $51.5 \pm 0.7$ & $\underline{93.0 \pm 0.9}$ & $81.8 \pm 0.9$ & $77.0 \pm 1.1$ & $\mathbf{71.7 \pm 1.9}$ & $\underline{61.9 \pm 2.3}$ & $\mathbf{76.6 \pm 4.5}$ & $\mathbf{36.5 \pm 1.3}$ & $\underline{83.8 \pm 4.7}$ & $83.7 \pm 3.7$ & $87.4 \pm 0.7$ & $78.7 \pm 0.7$ & $\underline{90.5 \pm 0.4}$ \\
Ordered GNN & $\underline{85.1 \pm 0.4}$ & $\mathbf{53.6 \pm 0.5}$ & $92.0 \pm 0.6$ & $\underline{84.6 \pm 0.5}$ & $\mathbf{77.9 \pm 1.2}$ & $\underline{71.2 \pm 1.5}$ & $60.1 \pm 2.3$ & $74.9 \pm 4.0$ & $35.1 \pm 1.3$ & $80.3 \pm 4.8$ & $86.1 \pm 3.5$ & $86.4 \pm 0.5$ & $\underline{78.8 \pm 1.0}$ & $\mathbf{90.6 \pm 0.3}$ \\
\textbf{CHAT-GNN} & $\mathbf{91.3 \pm 0.4}$ & $\underline{52.6 \pm 0.6}$ & $\mathbf{97.3 \pm 0.3}$ & $\mathbf{85.7 \pm 0.7}$ & $\underline{77.8 \pm 1.2}$ & $70.5 \pm 2.1$ & $\mathbf{64.3 \pm 1.5}$ & $72.2 \pm 4.2$ & $35.6 \pm 1.3$ & $\mathbf{84.3 \pm 5.4}$ & $\mathbf{87.3 \pm 3.2}$ & $87.0 \pm 0.5$ & $\mathbf{79.0 \pm 0.4}$ & $\mathbf{90.6 \pm 0.3}$ \\
\bottomrule
\end{tabular}

}
\label{tab:benchmark}
\end{table*}




\subsection{Implementation Details}
The models are trained with Adam optimization algorithm~\cite{adam}. The maximum number of epochs is set to 5000 with the early stopping patience of 500 epochs starting after 200 warm-up epochs. The early stopping is determined based on the validation accuracy. Weight decay is used to regularize the training. We tune the learning rate, weight decay rate, hidden dimension, number of layers, and model-specific hyperparameters using Bayesian optimization. We run 75 iterations of Bayesian optimization for each model and dataset. We evaluate the models using the average validation accuracy of the first three splits. Finally, we select the hyperparameter settings with the highest validation accuracy and re-train the models with the same configuration using the ten dataset splits.

The implementation of the proposed CHAT-GNN and baselines used in the experiments of this study with the best hyperparameters and their search spaces is publicly available at \footnote{\url{https://github.com/ALLab-Boun/CHAT-GNN}}. The official implementations of the recent baseline models, AERO-GNN \footnote{\url{https://github.com/syleeheal/AERO-GNN}} and Ordered GNN \footnote{\url{https://github.com/LUMIA-Group/OrderedGNN}} are used. We run both models with the reported best hyperparameters instructed by the authors for the common benchmark datasets. We use the implementations in the popular graph learning library called PyTorch Geometric~\cite{pyg} for other baseline models. When the performance on some of the benchmark datasets is missing in the baseline studies, we train the baseline models from scratch by running hyperparameter optimization considering the suggested hyperparameter setting by the authors, dataset scale, and our computational resources. We train all the models on a single GPU. We use NVIDIA Tesla T4 (16GiB) and NVIDIA Tesla V100 (16GiB) as our GPU hardware.




\subsection{Overall Performance}
Node classification performances of all baselines and CHAT-GNN are shown in Table~\ref{tab:benchmark}. The results in this table are obtained by training all the baseline models from scratch. The datasets minesweeper, tolokers, and questions by Platonov {\it et al.}~\cite{platonov} constitute binary classification problems. As the authors suggested, we report AUC for minesweeper, tolokers, and questions. For other datasets, we report mean classification accuracy. All datasets have ten train, validation, and test splits. We report the average performance with its standard deviation.

As seen in Table~\ref{tab:benchmark}, CHAT-GNN outperforms baseline methods on heterophilous graphs such as minesweeper, roman-empire, and squirrel. For citation networks with high homophily, CHAT-GNN is either comparable to the state-of-the-art or outperforms by a small margin. We can observe that the proposed model adapts to graphs of different characteristics in terms of domain, homophily, and size by achieving either superior or comparable performance with the well-known GNNs in the literature.


\subsection{Over-smoothing}

\subsubsection{Performance}
\begin{figure}[t]
    \centering
    \includegraphics[width=1\linewidth]{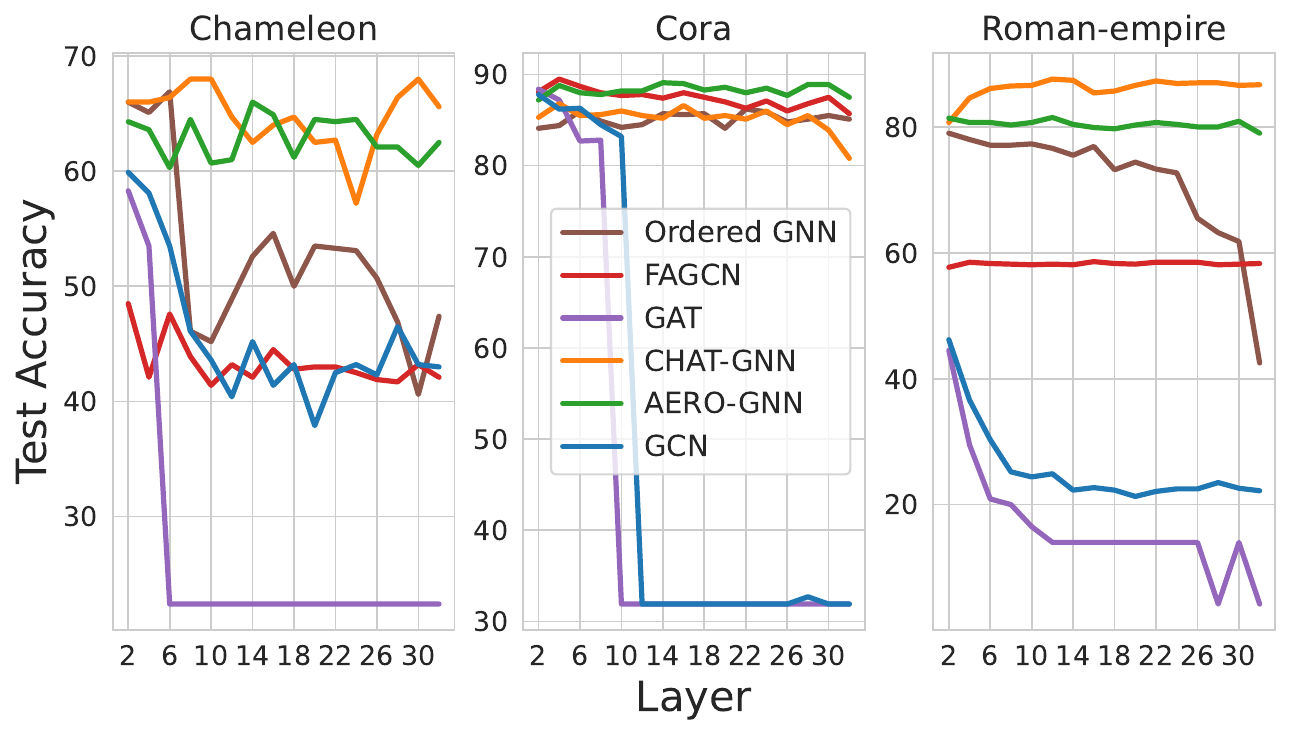}
    \caption{Resistance of selected models to over-smoothing. The change in test accuracy is plotted with respect to an increasing number of message-passing layers.  }
    \label{fig:layerwise_acc}
\end{figure}

A significant decrease in performance after a certain number of message-passing layers is a clear indication of over-smoothing in GNNs~\cite{oversmoothing, gnn_exp}. To investigate the over-smoothing in the proposed framework, we train CHAT-GNN and several message-passing GNNs with 2 to 32 layers on three datasets with differing characteristics. Figure~\ref{fig:layerwise_acc} shows the performance change with the increasing number of layers. We also use 32 as the number of hidden channels for all models without additional tuning. The results show that while standard models such as GCN and GAT suffer from over-smoothing at the early layers, the proposed model is able to mitigate over-smoothing and compete with AERO-GNN that combines edge attention and hop attention~\cite{aero}.


\subsubsection{Dirichlet Energy}
Dirichlet energy, introduced in~\eqref{eq:dirichlet}, calculates how much a signal $X$ changes under a graph structure and therefore serves as a measure of smoothness. In GraphCON study~\cite{graphcon}, the authors use the Dirichlet energy to quantify over-smoothing as the exponential decay of the Dirichlet energy of the hidden node features as we go deeper through the message-passing layers in a GNN~\cite{graphcon}.

\begin{figure}[t]
    \centering
    \includegraphics[width=1\linewidth]{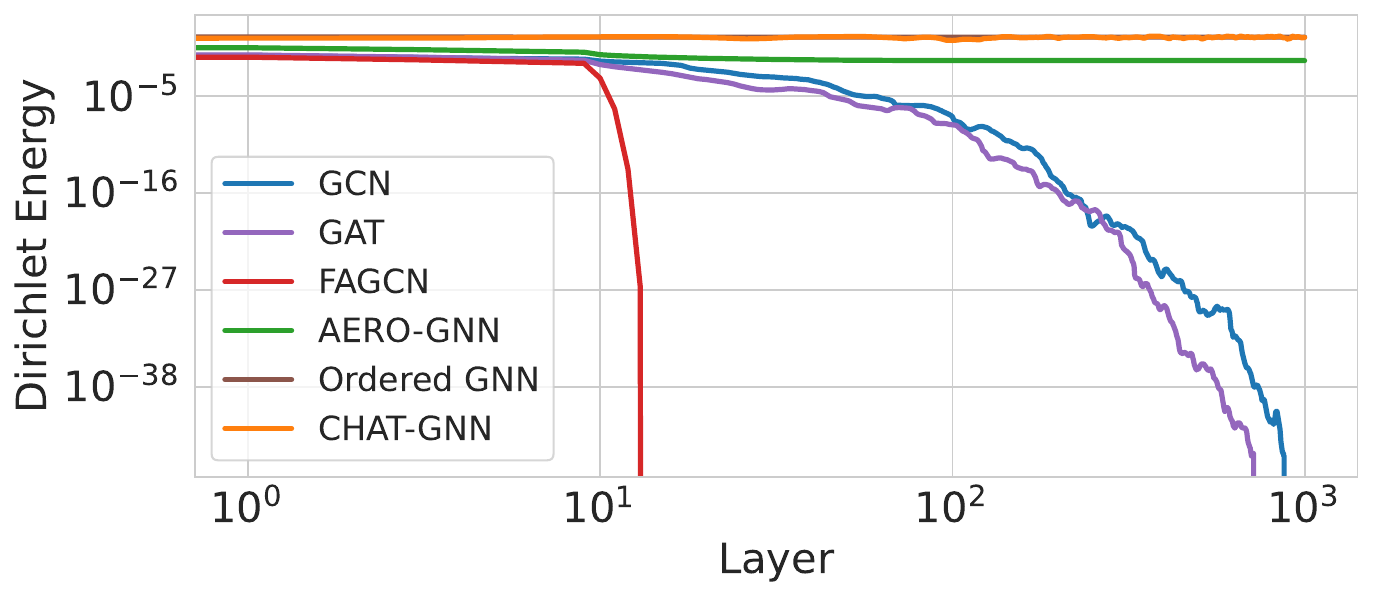}
    \caption{Change in Dirichlet energy of the message-passing layer output features of selected models with respect to increasing number of layers. }
    \label{fig:dirichlet}
\end{figure}

We use the same setting in GraphCON ~\cite{graphcon} to measure the Dirichlet energy of the output features of message-passing layers of various GNNs. Namely, we simulate message-passing on a $10 \times 10$ grid, in which each node has four neighbors. Each node has two features sampled from the uniform distribution, $\mathcal{U}(0,1)$. We run message-passing on layers with random weights up to 1000 layers. Figure~\ref{fig:dirichlet} shows how the Dirichlet energy changes with respect to the layers. The proposed CHAT-GNN keeps the Dirichlet energy almost constant and higher than AERO-GNN, while the baseline models reach almost zero Dirichlet energy in deeper layers, which indicates severe over-smoothing~\cite{graphcon}. The figure shows the scalability of CHAT-GNN to very deep message-passing layers.

\subsection{Visualizing Channel Weights}
\label{sub:vis}
\begin{figure}[t]
 \begin{minipage}[t]{0.49\linewidth}
    \centering
    \includegraphics[width=\linewidth]{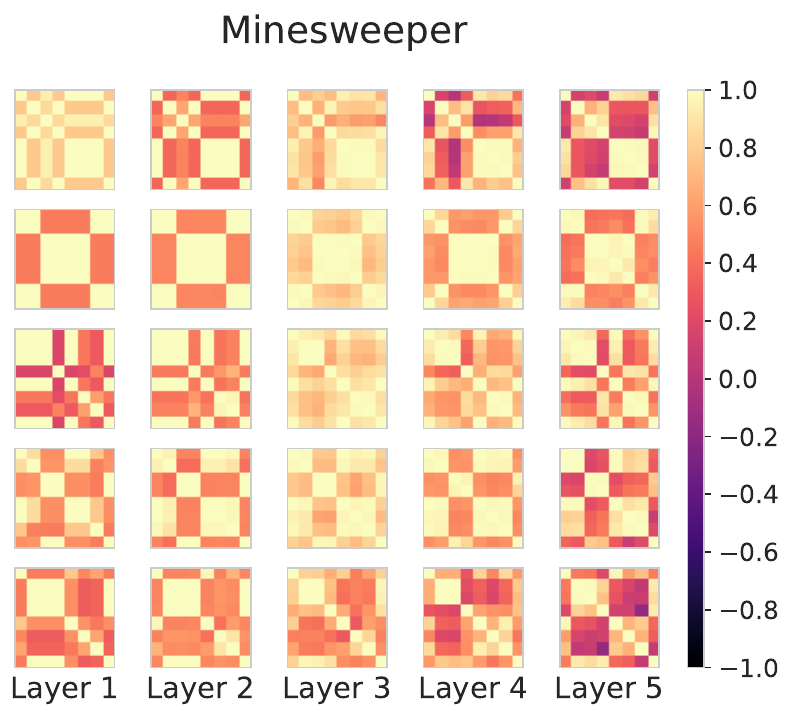}
  \end{minipage}
  \begin{minipage}[t]{0.49\linewidth}
    \centering
    \includegraphics[width=\linewidth]{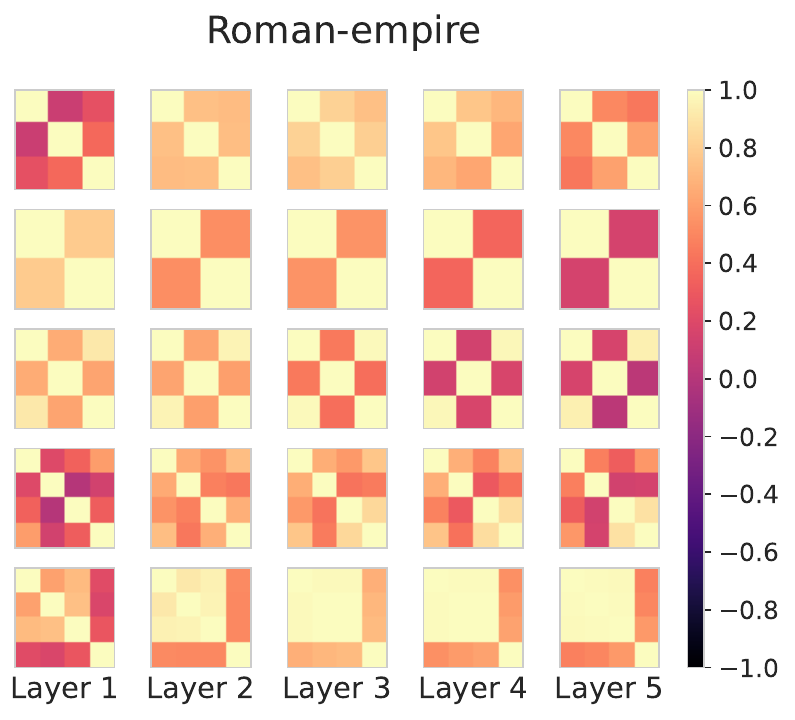}
  \end{minipage}
  \caption{Heatmap of pairwise cosine similarities between $\beta_{ji}$'s for a randomly selected node $i$ and its neighbor $j$ throughout the message-passing layers of CHAT-GNN.}
  \label{fig:heatmap}
\end{figure}

We visualize the weight vectors $\beta_{ji}$ through message-passing layers in CHAT-GNN in Figure~\ref{fig:heatmap}. As discussed in Section~\ref{sub:chat}, $\beta_{ji}$ denotes the channel-wise attention vector to evaluate the message function from node $i$ to node $j$. We visualize the pairwise cosine similarities, $ \frac{\beta_{ji}^T \beta_{ki}}{ \| \beta_{ji} \| \| \beta_{ki} \| } $, between each neighbor $i$ in Figure~\ref{fig:heatmap}. If the cosine similarity is close to one for an edge pair $(j, i)$ and $(k, i)$, the nodes $k$ and $j$ receive almost the same features from the node $i$. The results show that CHAT-GNN can learn to apply different message-passing schemes for different neighbors, nodes, and hops. Furthermore, while it sends diverse messages to neighbors in heterophilous graphs, it sends similar messages to neighbors in graphs with high homophily. However, if needed, it adjusts the message-passing mechanism to send more diverse messages. Note that we randomly draw five nodes from each dataset for illustration purposes. Also, we visualize the first five layers.

\subsection{Directed GNN Setting}
Edge directions are often not considered in the baseline GNNs. A message-passing layer is concerned with only the immediate neighbors of a node. It becomes a common practice to convert the graph to an undirected one to include both directions in message-passing. However, recent studies, such as Dir-GCN, suggest applying message-passing in both directions separately and with separate learnable parameters~\cite{dirgnn}. The bidirectional message-passing results show that when combined with Jumping Knowledge Networks~\cite{jknet} and a GCN or GAT layer as the base message-passing layer, the performance is boosted on some heterophilous graphs~\cite{dirgnn}. Therefore, we also report the bidirectional message-passing results of CHAT-GNN in Table~\ref{tab:dirgnn} on three heterophilous graphs. The results are obtained after hyperparameter tuning on Dir-GCN and Dir-CHAT-GNN. Table~\ref{tab:dirgnn} shows that adding edge directions improves the predictive performance of CHAT-GNN more than the standard GCN. Note that, unlike Dir-GCN, we achieve competitive results without employing Jumping Knowledge Networks in Dir-CHAT-GNN, which adds computational overhead.

\begin{table}[t!]
    \centering
    \caption{Performance comparison of CHAT-GNN and standard GCN under bidirectional message-passing setting.}
    \begin{tabular}{llll}
    \toprule
     & roman-empire & tolokers & chameleon \\
    \midrule
    Dir-GCN & $90.7 \pm 0.5$ & $77.7 \pm 1.1$ & $\mathbf{78.5 \pm 1.1}$ \\
    Dir-CHAT-GNN & $\mathbf{94.3 \pm 0.4}$ & $\mathbf{83.1 \pm 0.9}$ & $76.6 \pm 2.5$ \\
    \bottomrule
    \end{tabular}
    \label{tab:dirgnn}
\end{table}

\subsection{Ablation Study}
\label{sub:ablation}

\begin{figure}[t]
    \centering
    \includegraphics[width=1\linewidth]{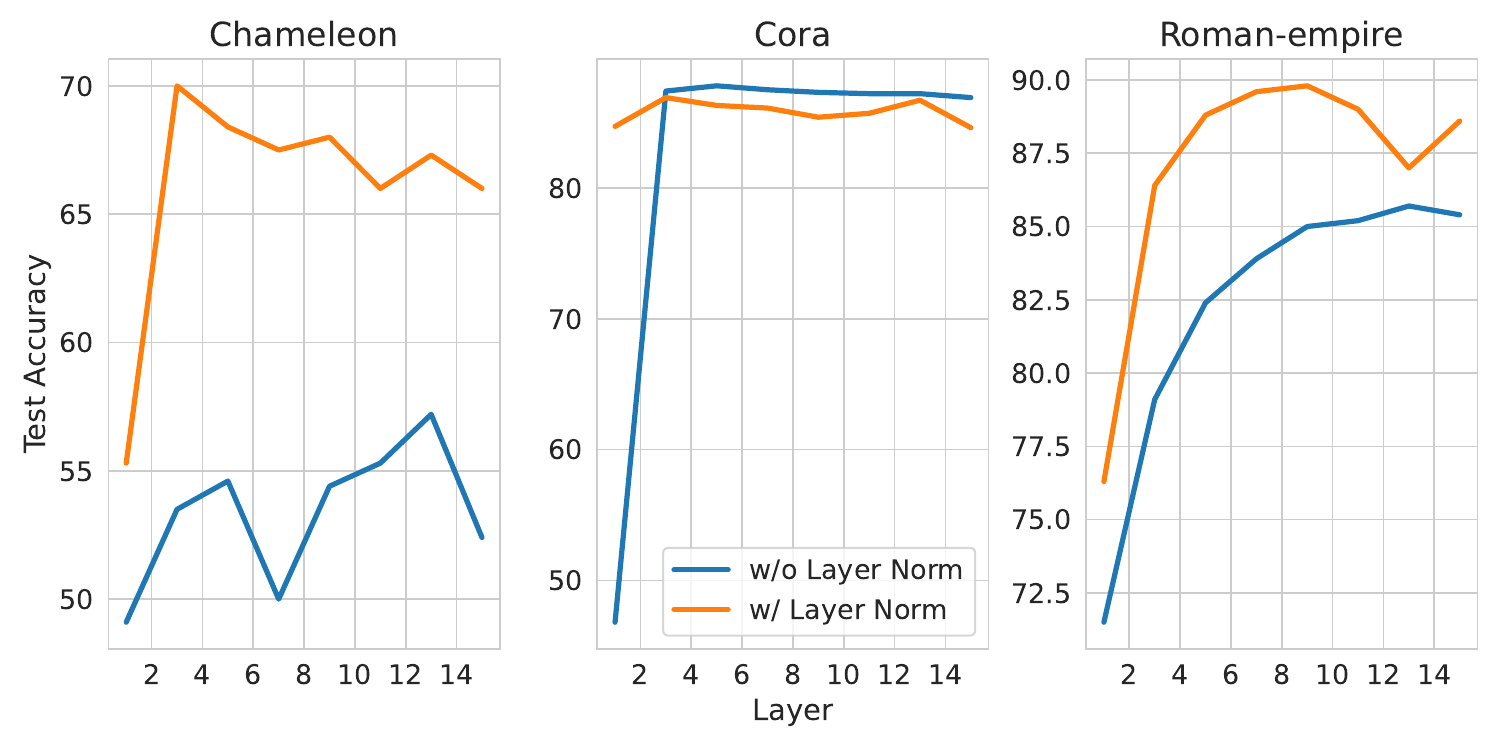}
    \caption{Effect of layer norm on the test accuracy as the number of layers increases.}
    \label{fig:layernorm_ablation}
\end{figure}
\begin{figure}[t]
    \centering
    \includegraphics[width=1\linewidth]{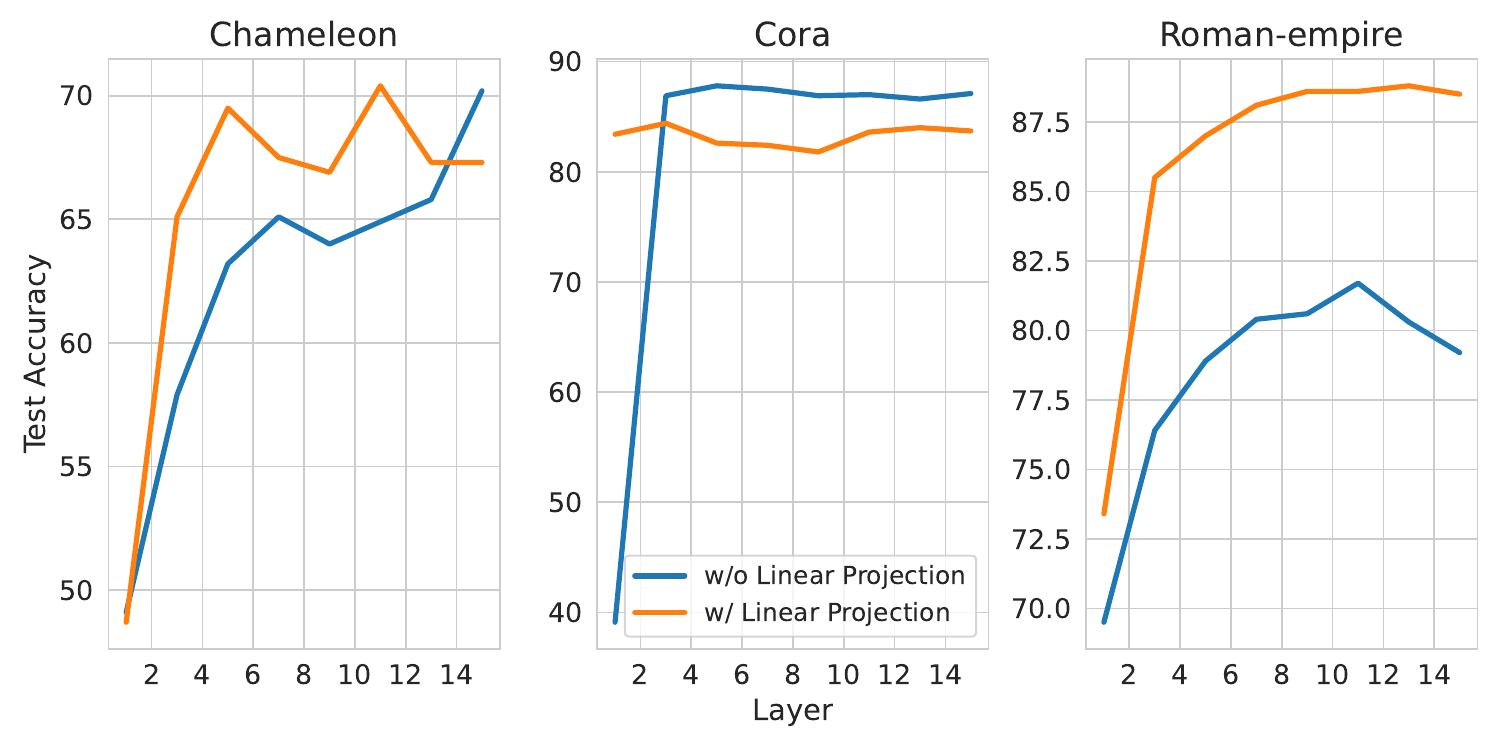}
    \caption{Effect of the separate learnable projection layers on the test accuracy as the number of layers increases.}
    \label{fig:projection_ablation}
\end{figure}
\begin{table}[t!]
    \centering
    \caption{The results when the channel-wise attention replaced by GAT and FAGCN layers, respectively.}
\begin{tabular}{llll}
\toprule
 & GAT & FAGCN & CHAT-GNN \\
\midrule
amazon-ratings & $36.9 \pm 0.1$ & $49.2 \pm 0.4$ & $\mathbf{52.6 \pm 0.6}$ \\
citeseer & $77.5 \pm 0.1$ & $77.5 \pm 0.0$ & $\mathbf{79.0 \pm 0.4}$ \\
minesweeper & $92.6 \pm 6.8$ & $95.4 \pm 0.3$ & $\mathbf{97.3 \pm 0.3}$ \\
roman-empire & $78.0 \pm 2.4$ & $83.5 \pm 0.7$ & $\mathbf{91.3 \pm 0.4}$ \\
squirrel & $24.4 \pm 9.0$ & $51.1 \pm 2.7$ & $\mathbf{64.3 \pm 1.5}$ \\
tolokers & $72.3 \pm 3.8$ & $84.0 \pm 1.1$ & $\mathbf{85.7 \pm 0.7}$ \\
\bottomrule
\end{tabular}
\label{tab:chat_ablation}
\end{table}

We conduct an ablation study on layer normalization and separate learnable projection modules of CHAT-GNN. Instead of a setting with a fixed number of layers, we train CHAT-GNN models with varying numbers of message-passing layers and visualize the effect of a specific component. 

\subsubsection{Layer Normalization}
We observe that layer normalization usually has a positive effect on classification performance, as seen in Figure~\ref{fig:layernorm_ablation}. It rescales the combined node and neighborhood representation, and intuitively, this may help the optimization process. For the citation network dataset Cora, we observe that adding layer normalization slightly reduces the accuracy. This is because the citation network datasets Cora, Citeseer, and Pubmed can easily be overfitted, as observed in all the baselines.

\subsubsection{Linear Projection}
We investigate the effect of adding separate learnable linear projection layers to the node and the neighborhood representations in Figure~\ref{fig:projection_ablation}. For Chameleon, we observe that the model without projection layers eventually surpasses the models with projection layers. In Cora, we see an effect similar to layer normalization. For the roman-empire, we see a substantial improvement with linear projection layers.
\subsubsection{Channel-wise Attention}
Finally, we conducted another ablation study that measures the effect of channel-wise attention. We use the architecture described in Section~\ref{sec:whole} and replace the channel-wise attention layer with GAT~\cite{gat} and FAGCN~\cite{fagcn} layers. The results in Table~\ref{tab:chat_ablation} show that channel-wise attention indeed contributes to the state-of-the-art results.
\section{Conclusion}
This study proposes a message-passing GNN architecture that can attend to the feature channels of each neighbor. Experiments show its superiority and ability to alleviate over-smoothing. We also visualize the learned attention vectors that are used in channel-wise message-passing. The illustrations show that the proposed model can adjust the weights according to the node's neighbors and hops. In future work, the scalability of the CHAT-GNN will be investigated for applying it to large-scale benchmarks such as Open Graph Benchmark~\cite{ogb}. The proposed channel-wise attention mechanism could stand out even more in large-scale settings where longer-range relationships must be captured. In addition, graph-level inductive tasks will also be investigated to see how the proposed model performs in learning graph-level representations on unseen graphs.


\bibliographystyle{IEEEtran}
\bibliography{main}

\end{document}